\documentclass[letterpaper, 10 pt, conference]{ieeeconf}  

\IEEEoverridecommandlockouts                              

\usepackage{amsmath}
\usepackage{amssymb}
\usepackage{todonotes}
\usepackage{booktabs}
\usepackage{array}
\usepackage{soul, color}
\usepackage{float} 
\usepackage{subcaption}
\usepackage{subfig}
\usepackage{multirow}
\newcommand{\website}{https://zhuoheng0910.github.io/dribble-master/}
\usepackage{caption}
\usepackage{url}

\title{\LARGE \bf
Dribble Master: Learning Agile Humanoid Dribbling 

through Legged Locomotion
}

\author{Zhuoheng Wang$^{\dagger}$$^{,}$$^{*}$, Jinyin Zhou$^{\dagger}$$^{,}$$^{*}$, Qi Wu$^{\ddagger}$$^{,}$$^{1}$\\
$^{\dagger}$Tsinghua University \quad $^{\ddagger}$Cornell University \quad $^{*}$Equal Contribution\\
}

\begin{document}
\twocolumn[{%
\renewcommand\twocolumn[1][]{#1}%
\maketitle
\begin{center}
    \vspace{-0.1in}
    \centering
    \captionsetup{type=figure}
    \includegraphics[width=\textwidth]{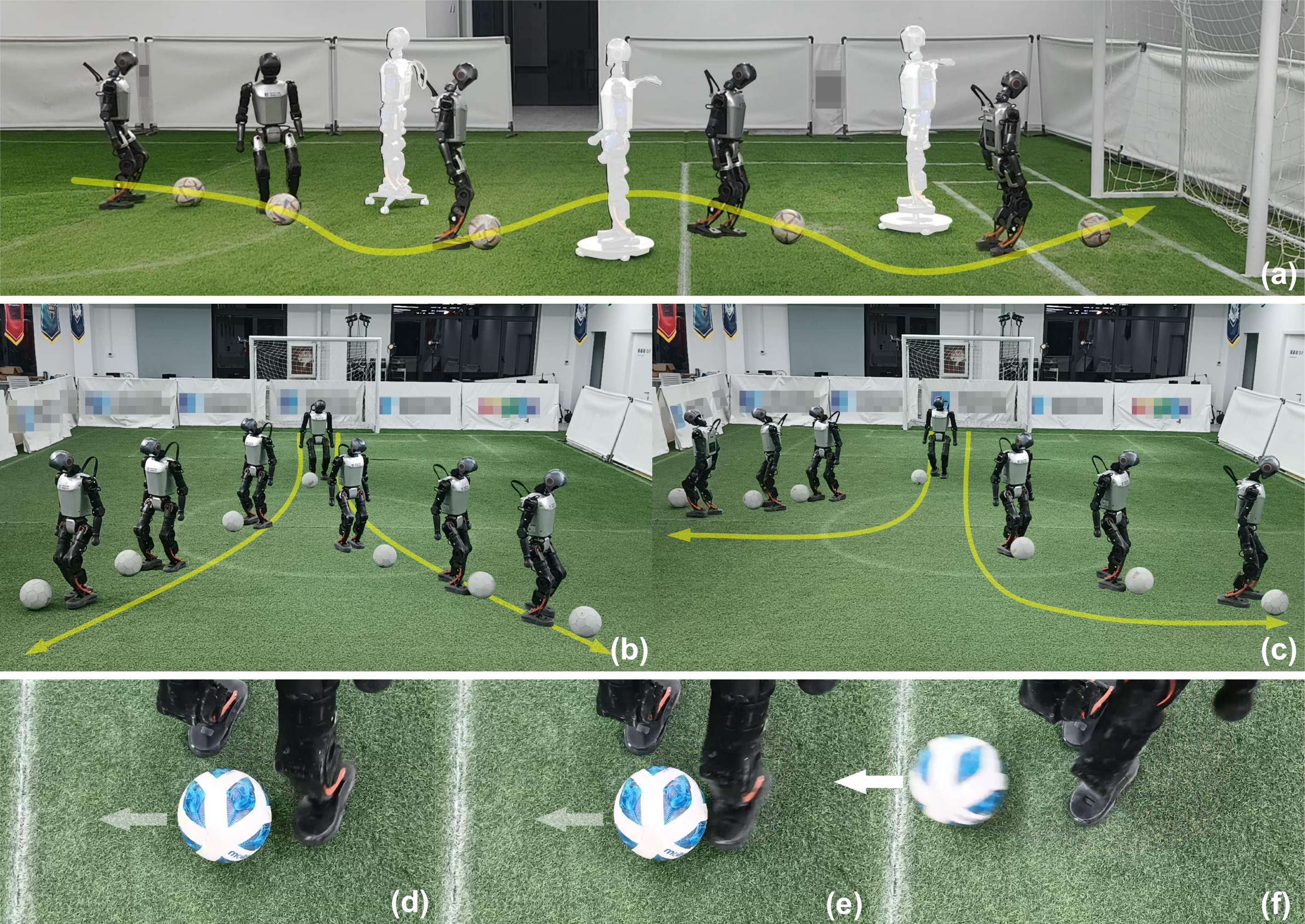}
    \captionof{figure}{\textbf{Dribble Master: Humanoid robot learning to dribble under various tasks.} \textbf{(a):} The robot receives ball velocity commands via a joystick and successfully navigates an array of three obstacles, ultimately dribbling the ball to the goal with high agility and precision. \textbf{(b)(c):} The robot exhibits agile directional-change capabilities. (b) combining straight-line movement with left-front and right-front turns; (c) combining straight-line movement with sharp left and right turns. \textbf{(d)(e)(f):} Detailed motion sequence showing the robot executing a rightward direction change by making contact with the ball.}
    \label{fig:cover}
\end{center}%
}]

\thispagestyle{empty}
\pagestyle{empty}

\newcommand{\ourmethod}{Our Method}


\footnotetext[1]{This work was done before the author joined Cornell University.}

\begin{abstract}

Humanoid soccer dribbling is a highly challenging task that demands dexterous ball manipulation while maintaining dynamic balance. Traditional rule-based methods often struggle to achieve accurate ball control due to their reliance on fixed walking patterns and limited adaptability to real-time ball dynamics. To address these challenges, we propose a two-stage curriculum learning framework that enables a humanoid robot to acquire dribbling skills without explicit dynamics or predefined trajectories. In the first stage, the robot learns basic locomotion skills; in the second stage, we fine-tune the policy for agile dribbling maneuvers. We further introduce a virtual camera model in simulation that simulates the field of view and perception constraints of the real robot, enabling realistic ball perception during training. We also design heuristic rewards to encourage active sensing, promoting a broader visual range for continuous ball perception. The policy is trained in simulation and successfully transferred to a physical humanoid robot. Experiment results demonstrate that our method enables effective ball manipulation, achieving flexible and visually appealing dribbling behaviors across multiple environments. This work highlights the potential of reinforcement learning in developing agile humanoid soccer robots. Additional details and videos are available at \website.

\end{abstract}

\section{INTRODUCTION}

Learning-based locomotion has recently enabled humanoid robots to run, crawl, and even dance. However, to make humanoids practical in real-world tasks, it is crucial to develop their ability to interact with objects and accomplish complex tasks. Current research on robot-object interaction has focused primarily on arm-based mobile manipulation. Recent works \cite{fu2024humanplus, he2024omnih2o} have shown that humanoid robots can complete complex manipulation tasks such as folding clothes and boxing. However, little attention has been given to loco-manipulation that involves rich contacts using the legs.

Humanoid soccer dribbling presents a challenging example of legged loco-manipulation. First, it is difficult to model the dynamics of the robot and the interactions between the ball and the robot's feet. During dribbling, the robot must subtly manipulate the ball to control its velocity and direction, while simultaneously maintaining dynamic balance. Second, modeling the rolling dynamics of the ball across diverse terrains — ranging from artificial grass to potholed ground, with varying surface properties like friction and compliance — further complicates controller design, especially for methods that require accurate models of such interactions \cite{chiu2022collision}. Third, discontinuous visual perception poses a significant challenge. While the use of wide-angle fish-eye cameras \cite{ji2023dribblebot} can expand the robot's field of view and improve ball observability, occlusions and distortions often degrade ball localization.

Several prior efforts have explored humanoid soccer dribbling. Many methods developed in RoboCup Soccer Competitions employ modularized methods, using walking gaits to interact with the ball. However, these methods rely on incidental foot-ball contacts during gait cycles, limiting the robot’s ability to actively control ball direction. More recent research \cite{tirumala2024soccervision, haarnoja2024learning} has tried end-to-end learning, but often at the cost of generalization across environments. Hierarchical methods \cite{ji2023dribblebot, peng2017deeploco} have been explored but struggle with reliable ball tracking and continuous control.

In our work, we present Dribble Master, a system that enables a humanoid robot to perform agile and robust soccer dribbling skills across diverse environments. We employ a two-stage curriculum learning framework, introduce a virtual camera model in simulation, and incorporate active sensing in our training pipeline. We train our dribbling policy in simulation and successfully transfer it to the physical Booster T1 humanoid robot. Experiment results demonstrate that our method achieves great agility, dexterity, and robustness in challenging dribbling tasks, outperforming traditional approaches in terms of success rate, efficiency, and visual appeal. We believe our work can help promote further research on legged loco-manipulation, robot soccer, and athletic intelligence for humanoid robots.

Our key contributions can be summarized as follows:
\begin{itemize}
    \item We propose a two-stage curriculum learning framework with the corresponding reward system for humanoid dribbling, enabling the robot to gradually acquire dribbling skill while avoiding sparse rewards and early terminations caused by unstable and asymmetric contacts.
    \item We introduce a virtual camera model in simulation and design heuristic reward functions for vision-based ball seeking and tracking. Our robot learns to actively sense the ball, which expands the field of view and improves robustness against discontinuous visual perception.
    \item We successfully deploy our dribbling policy onto the physical humanoid robot Booster T1. In the real world, our method has demonstrated agility and precision in real-world dribbling under different challenging scenarios. To the best of our knowledge, this is the first demonstration of learning-based humanoid soccer dribbling across multiple terrains in the real world.
\end{itemize}

\section{RELATED WORK}
\subsection{Dynamic Locomotion on Humanoid Robots with RL}
Reinforcement learning (RL) has shown impressive results in humanoid locomotion. Recent works have demonstrated diverse and agile locomotion skills using model-free RL \cite{gu2024humanoid, gu2024advancing, li2024reinforcement}. Multi-stage training schemes have been explored to improve robustness in tasks such as jumping and rapid recovery \cite{li2023robust, chen2025hifar}. Some approaches have also integrated vision to navigate complex terrains \cite{duan2024learning}.
However, these studies mainly address locomotion, while our approach tackles dynamic dribbling, which demands both stable locomotion and precise object interaction.

\subsection{Loco-Manipulation on Legged Robots}
Prior work has enabled quadruped manipulation using dedicated arms \cite{sleiman2021unified, chiu2022collision, fu2023deep, ha2024umi} or unconventional tools \cite{wu2024helpful}, increasing mechanical complexity. Other methods repurpose legs for manipulation: e.g., pressing buttons \cite{cheng2023legs}, manipulating a circus ball \cite{shi2021circus}, or using loco-manipulators for dexterous control \cite{lin2024locoman, he2024learning, arm2024pedipulate}. While these works show promise, they mainly target quadruped platforms. 
Recently, whole-body motion control for humanoid robots has become a focal point of research. Several emerging works \cite{falcon2025,xue2025hugwbc,exbody2_2024,sun2025spark,lu2024pmp,harmon2025} tackle the coupled problem of locomotion and manipulation by integrating upper-body dexterity with legged mobility, yet they typically emphasize either precise hand-object interaction or expressive upper-body gestures.  In parallel, another line of studies \cite{liu2024opt2skill,he2025asap,he2025gettingup,Wang2025skillmimic} focuses on replicating agile, whole-body maneuvers such as basketball shots or stand-up motions. However, research on humanoid loco-manipulation, particularly using feet to interact with dynamic objects, remains scarce.

\subsection{Legged Robot Soccer Skills}
Robot soccer has long been a benchmark task for legged locomotion and coordination, most notably through the RoboCup competition series \cite{kitano1997robocup}. While early RoboCup efforts focused on rule-based or scripted control, recent years have seen a growing interest in learning-based methods \cite{gerndt2015humanoid, ji2022hierarchical, huang2023creating}. However, these methods are basically implemented on small robots such as Nao and cannot achieve continuous ball touching and dribbling \cite{Leottau2015fuzzy}. Quadruped robots have achieved impressive dribbling and goalkeeping skills using RL \cite{ji2023dribblebot}, while bipedal approaches have mostly been limited to discrete kicking \cite{peng2017deeploco}, which lacks continuous control over the ball.
Humanoids face unique challenges: a smaller support polygon and slower balance recovery make continuous dribbling significantly harder than in quadrupeds. Recent end-to-end methods for humanoid soccer skills have achieved agile control in simulation \cite{tirumala2024soccervision, haarnoja2024learning}, but often rely on specific, structured environments.
While previous works have demonstrated impressive results in ball manipulation with quadrupeds and kid-sized humanoids, agile and terrain-generalized dribbling for full-sized humanoids remains underexplored. In contrast, our method enables a bipedal humanoid to perform refined, multi-touch dribbling in diverse and unstructured environments, pushing the boundary of loco-manipulation on humanoid platforms.

\begin{figure*}[h]
    \centering
    \includegraphics[width=1\linewidth]{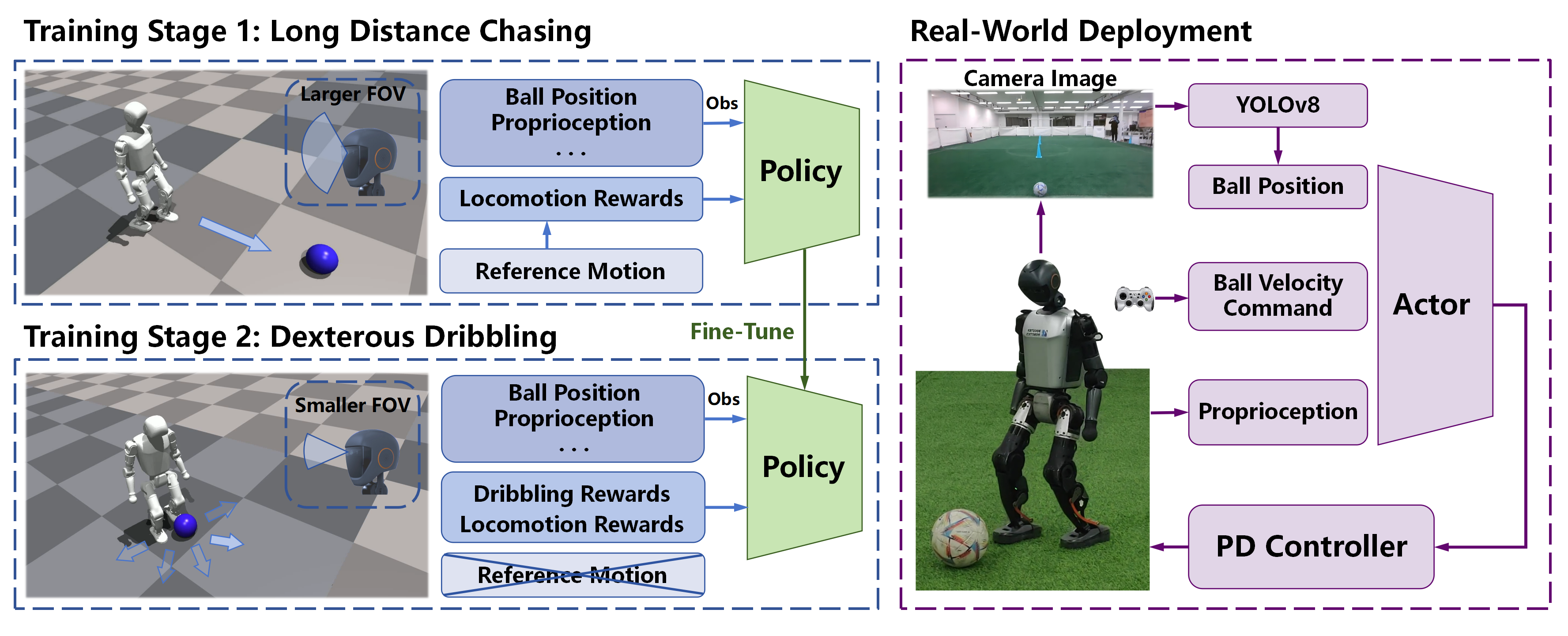}
    \caption{\textbf{System Architecture of Dribble Master}. In the phase of training in simulation, we use a two-stage learning approach. During the 1st stage, the locomotion rewards are given higher weights, with the ball far away from the robot. This aims to train the robot to run to the ball stably and rapidly. During the 2nd stage, the balls are near the robot and the dribbling rewards have non-zero high weights, which enables the robot to learn to manipulate the ball for precise velocity tracking. In the phase of real-world deployment, we transfer the trained actor policy to a physical Booster T1 humanoid robot. The ball position is obtained via the YOLOv8 module for ball detection of raw images. The target ball velocity is commanded by a joystick maneuvered by humans.}
    \label{fig:method_pic}
\end{figure*}

\section{Methods}
\subsection{Reinforcement Learning Problem Formulation}

We model humanoid dribbling as a Markov Decision Process (MDP) ($S$, $A$, $f$, $r_t$, $S_0$, $\gamma$), where $S$ is the state space, $A$ is the action space, $f$ is the system dynamics, $r_t$ is the reward function, $S_0$ is the initial state, and $\gamma$ is the discount factor. The objective of Reinforcement Learning (RL) is to learn a policy $\pi$ that maximizes the expected discounted return $G=E_\pi[\Sigma_{t=0}^{T-1}\gamma^tr_t]$.
We adopt Proximal Policy Optimization (PPO) \cite{schulman2017proximal} with an asymmetric actor-critic architecture and use privileged information during training, while relying on partial observations at deployment. 

\subsubsection{Observation Space}
Our observation space consists of three parts: commands, proprioception, and clock signals.

\textbf{Commands}. Previous work mainly uses robot velocity \cite{gu2024humanoid}, which is not intuitive for the ball dribbling task. We adopt the $x$ and $y$ ball velocity as the commands directly, under the assumption that the ball almost always stays on the ground. We adopt ball velocity tracking as the control objective to facilitate downstream tasks that require precise velocity modulation, such as fast dribbling or goal shooting. The robot learns to place its foot and adjust its kicking force to align with the magnitude and direction of the target velocity. The ball's velocity command $\mathbf{v}^{cmd}$ is described in the global frame. 

\textbf{Proprioception.} The policy takes in robot states including joint positions $q$, joint velocities $\dot{q}$ and body orientation $\theta_{yaw}, \theta_{roll}, \theta_{pitch}$ as inputs. Inspired by \cite{ji2023dribblebot}, we use ball position $\mathbf{b}$ instead of raw images as input for ball perception. We further include an indicator of whether the ball is in the field of view to support active sensing. 

\textbf{Clock Signal}. To accelerate the training process, we use a sinusoidal clock signal \cite{gu2024humanoid} as a pattern generator to encourage rhythmic leg motion and natural gaits. Specifically, we introduce a periodic phase variable into the observation space. The sine and negative sine of this variable are used as simple reference signals for the hip pitch joints, serving as a guide to promote rhythmic gait patterns during learning.

\subsubsection{Action Space}
The action space has 14 dimensions, controlling joint target positions for the 2-DOF head and 12-DOF legs. These targets are tracked by PD controllers, which generate torques to the lower-level motor control.

\subsubsection{Reward Functions}
We adopt a two-stage training pipeline with different reward functions and weights in each stage. The goal is to first learn the basic locomotion skill and then learn the more fine-grained ball manipulation skills. Our discount factor $\gamma$ is 0.994. More details are discussed below.

\subsection{Two-Stage Curriculum Learning}
Training a humanoid to dribble presents a fundamental challenge: the rewards for walking and dribbling often conflict and are difficult to balance. Walking requires the robot to maintain stability and move efficiently, while dribbling demands agility and precise ball control. This trade-off complicates the simultaneous optimization of both skills.

Inspired by vision-language-action (VLA) research, such as RT-2 \cite{brohan2023rt}, which pretrains a general model before fine-tuning on specific tasks, we adopt a two-stage learning strategy for humanoid robot dribbling (Fig. \ref{fig:method_pic}). Our approach decouples the learning of low-level locomotion skills from ball manipulation, enabling more stable and efficient training.

In the first stage, the robot is trained to acquire fundamental locomotion skills, including stable walking and efficient ball chasing. To prioritize these abilities, we assign higher weights to locomotion rewards (e.g., reward for reference tracking and penalties for the torso tilt and stepping cycle mismatches) and ball-chasing rewards (e.g., penalty for the distance to the ball), while removing ball velocity tracking rewards. We initialize the ball far away from the robot to focus the learning process on the robot locomotion. This setup encourages the robot to stay upright and approach the ball actively, resulting in dense and consistent reward signals.

Once the robot consistently demonstrates reliable locomotion and target pursuit, we manually transition to the second stage, which focuses on fine-grained ball manipulation. We increase the weights of dribbling rewards (e.g., penalty for mismatched projected ball velocity) and initialize the ball closer to the robot. Reference motion rewards are removed to promote exploration and discovery of effective dribbling behaviors. This curriculum design mitigates early failures caused by poor ball-approach behavior and facilitates the emergence of nuanced ball control.

By structuring training into these two distinct phases, we effectively balance the competing objectives of stable locomotion and precise dribbling. This curriculum accelerates convergence and yields a robust policy capable of executing agile, robust behaviors in real-world scenarios. 

\subsection{Virtual Camera Model in Simulation}
\label{sec:active}
Previous work on quadrupedal dribbling \cite{ji2023dribblebot} employs a fisheye camera to achieve a wide field of view (FOV) for ball tracking. However, humanoid dribbling requires much more precise ball localization. Severe image distortion from fish-eye lenses requires exact calibration, which is prone to errors and unreliable in real-world settings. Additionally, rectangular cameras suffer from a narrow field of view, making it easy for the ball to leave the robot's vision during dribbling. 
To address these challenges, we propose training the robot to learn active sensing by tactically moving its head to seek the ball. In simulation, we implement a virtual camera model to support the emergence of this behavior.

\begin{figure}[!htbp]
    \centering
    \includegraphics[width=1\linewidth]{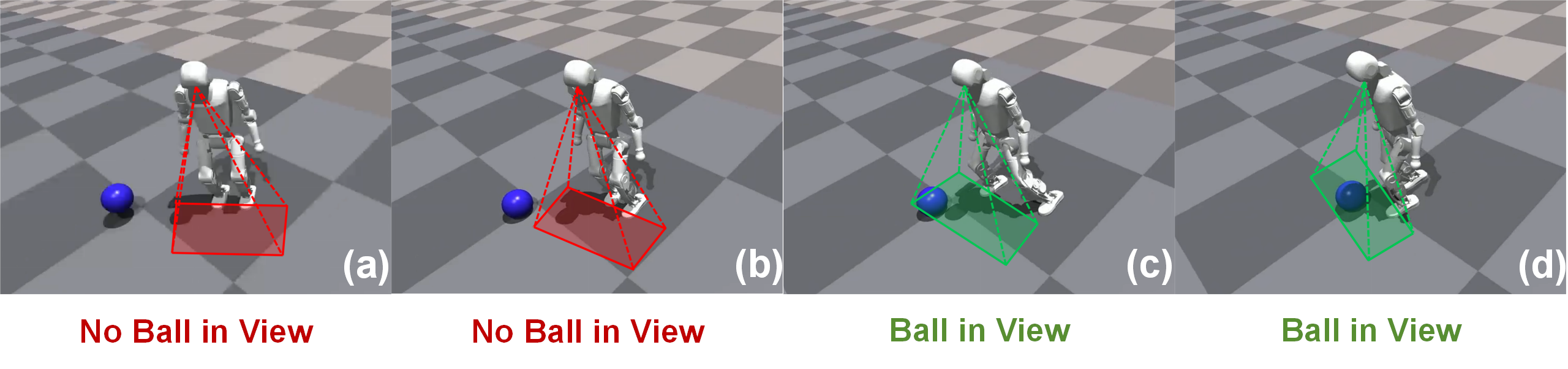}
    \caption{Active sensing rewards encourage the robot to search for the ball. When the ball is not in the view((a) and (b)), the robot has a smaller reward than (c) and (d).}
    \label{fig:search}
\end{figure}
As shown in Fig. \ref{fig:search}, we simulate a virtual camera model based on the RealSense D455 parameters: horizontal field of view (HFOV) and vertical field of view (VFOV), aligned with the hardware used on our Booster T1 robot. Since we have a simulated camera model with specified field of view and delay characteristics, we can determine whether the ball is visible and, if so, extract its position within the image plane. When the ball moves outside the virtual camera's field of view, we stop updating the ball's position in the policy's observation. A rule-based reward is provided when the ball is within view, encouraging the robot to maintain visual contact. As a result, the robot learns to align its vision with the ball when visible and to rotate its head and body to search for the ball when it is not. We maintain a 0.3-second history of the ball’s location, allowing the robot to continue moving toward the last known direction. If the ball is lost beyond this period, the robot learns to initiate rotation toward that direction to reacquire visual contact.

To improve learning speed and efficiency, we integrate this virtual camera model with our two-stage curriculum learning framework by having different camera parameters. In the first training phase, the camera’s field of view is artificially enlarged—set to twice the D455's actual FOV—allowing the robot to detect the ball more easily and receive dense rewards. In the second stage, the FOV is reduced to match the actual RealSense D455, promoting precise visual tracking and alignment with real-world deployment. This virtual camera module enables the robot to acquire effective active sensing capabilities through policy learning, which translates well to real-world performance.

\subsection{Training Environment in Simulation}
\subsubsection{Simulation Setup}
We train the policy using Humanoid Gym RL framework \cite{gu2024humanoid} in Isaac Gym \cite{makoviychuk2021isaac} and evaluate it in MuJoCo and Webots. The simulated robot is based on the URDF model of the Booster T1 humanoid robot,
with 23 DOFs. For simplicity,  we control 12 leg joints and 2 head joints, fixing the upper body. 
Training is parallelized with 4,096 agents for efficiency.
During training, the ball is initialized 10 meters from the robot in the first stage and within 2 meters in the second stage, with its target velocity updated every 4 seconds to promote dynamic interaction.

\subsubsection{Policy Architecture}
The actor network is represented by a Multi-Layer Perceptron (MLP) with the hidden layer sizes of [512, 256, 128]. The critic network, also an MLP with hidden layers [768, 256, 128], receives additional privilege state information directly from the simulation environment to enable more accurate value estimation.

\subsubsection{Measures to Bridge the Sim-Real Gap}
To bridge the sim-to-real gap and enhance the real-world performance of the dribbling policy, we have taken several techniques without the loss of agility and motion precision.

\begin{itemize}
    \item \textbf{Simulating Motor Delay:} We simulate motor latency by applying random actuation delays of 0-20 milliseconds, reflecting real hardware communication and computation delays. Joint PD controllers then compute actual torques based on the policy’s target positions.
    \item \textbf{Training on Rough Terrains:} To enhance the stability of dribbling on rough terrains, we have trained the robot to walk and dribble on terrains with various slopes, friction, and elevation noises.
    \item \textbf{Simulating Ball Perception Inaccuracy:} To address visual perception noises in the real world, we inject additional noise into the observed ball position.
    \item \textbf{Domain Randomization:} We apply domain randomization to physical parameters
     (see Table~\ref{tab:domain_randomization}), sampling each from a uniform distribution to enhance policy robustness and generalization.
\end{itemize}

\begin{table}[h]
\centering
\begin{tabular}{lllll}
\toprule
\textbf{Parameter} & \textbf{Range} & \textbf{Unit}  \\
\midrule
Action Scale & [0.95, 1.05] & - \\
Terrain Friction & [0.5, 1.5] & - \\
Base Mass & [-2, 2] & kg \\
Base CoM Position & [-0.04, 0.04] & m \\
Joints Kp Scale& [0.7, 1.3] & - \\
Joints Kd Scale& [0.8, 1.2] & -   \\
Joints Torque Scale& [0.95, 1.00] &  - \\
Joints Position & [-0.02, 0.02] & rad  \\
\bottomrule
\end{tabular}
\caption{Key Domain Randomization Parameters.}
\label{tab:domain_randomization}
\end{table}

\begin{table*}[h]
    \begin{center}
    \begin{tabular}{ccccc}
    \toprule
     & \multicolumn{2}{c|}{Dribbling to Target} & \multicolumn{2}{c}{Obstacle Avoidance} \\
    \cmidrule(lr){2-3} \cmidrule(lr){4-5}
    Method & Success Rate(\%) & Time Usage(s) & Success Rate(\%) & Time Usage(s) \\
    \midrule
    Direct Tele-Operation (Default Controller) & 66.7 & 42.3 & 60.0 & \textbf{23.0} \\
    \ourmethod{} without Active Sensing & 13.3 & \textbf{22.0} & 0 & NaN \\
    Single Stage Training & 33.3 & 29.7 & 0 & NaN \\
    \textbf{\ourmethod{}} & \textbf{86.7} & \textbf{22.0} & \textbf{93.3} & 31.4 \\
    \bottomrule
    \end{tabular}
    \end{center}
    \caption{We compared our method with 3 baselines. Our method has the highest success rate in both tasks and has a relatively short completion time.}
    \label{table:results}
\end{table*}

\subsection{Real-World Deployment}
\subsubsection{Hardware System}
We deploy our dribbling policy on the physical Booster T1 humanoid robot, a 1.18-meter-tall robot with 23 DOFs, an IMU, joint encoders for real-time state estimation, and a RealSense D455 Depth Camera for visual perception. Onboard computation uses an NVIDIA AGX Orin GPU and a 14-core high-performance CPU, running the dribbling policy at 50 Hz.

\subsubsection{Object Detection}
For real-time ball perception in the real world, we integrate the YOLOv8 model \cite{varghese2024yolov8} into the robot's vision pipeline. YOLOv8 detects the ball from RGB images and outputs bounding box coordinates, which are transformed into the robot’s base frame via a pre-calibrated camera-to-robot extrinsic transformation for accurate position estimation. The detection module runs at 30 Hz, accelerated by TensorRT for low-latency inference.
\subsubsection{Joystick Command}
A human operator provides ball velocity commands through a joystick interface. The joystick’s displacement is mapped to a 2D velocity vector $(v_x, v_y)$ in the global frame, allowing the operator to specify the desired direction and magnitude of the ball’s motion in real time. This command interface enables intuitive control over dribbling behavior during testing and demonstrations.

\section{EXPERIMENTS AND RESULTS}
\subsection{Velocity Tracking Accuracy in Simulation}
\subsubsection{Experiment Setup}

\begin{figure}[h]
    \centering
    \begin{subfigure}[b]{0.8\linewidth}
        \includegraphics[width=\linewidth]{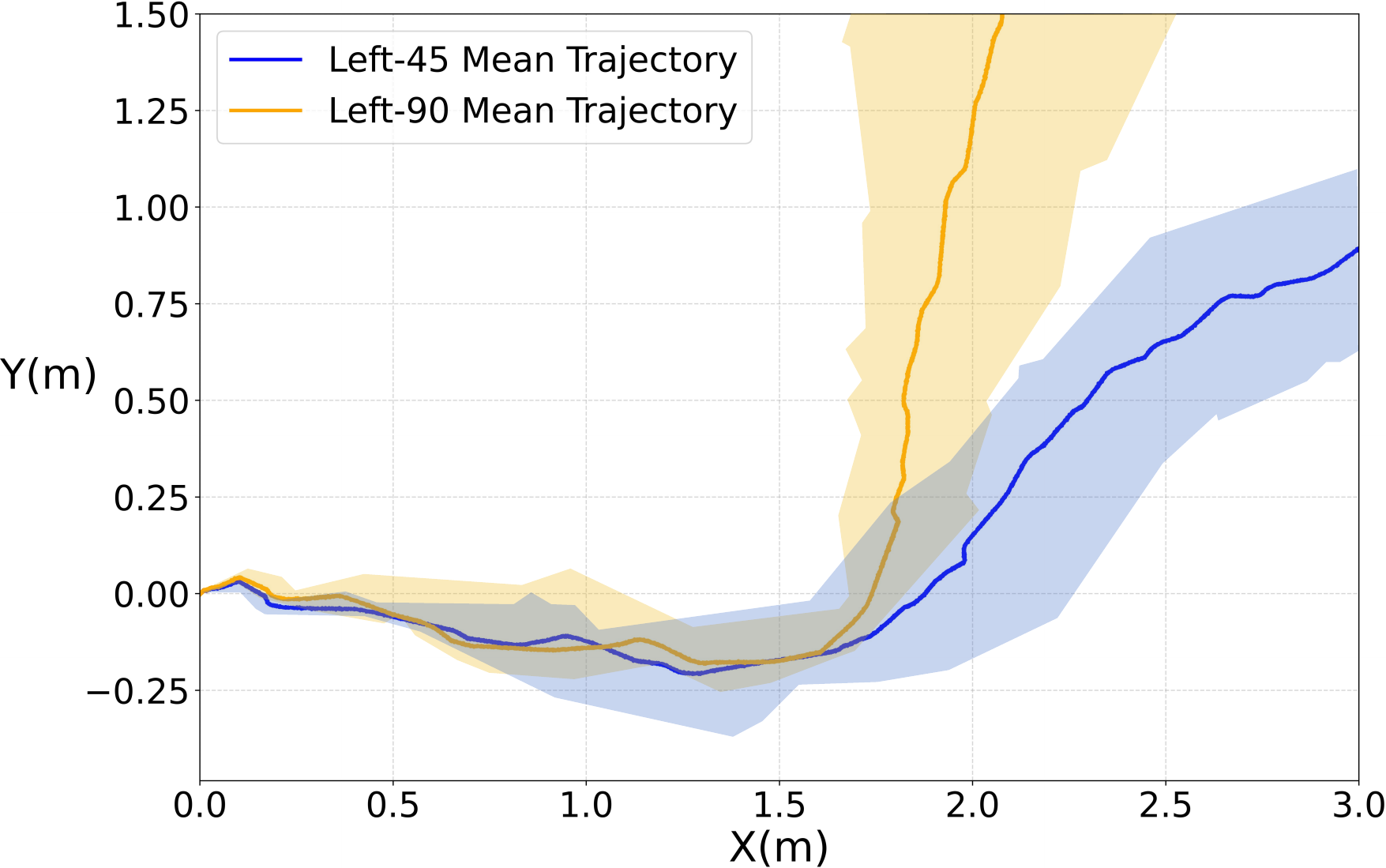}
        \caption{Left-turn dribbling trajectory}
    \end{subfigure}
    \hfill
    \begin{subfigure}[b]{0.8\linewidth}
        \includegraphics[width=\linewidth]{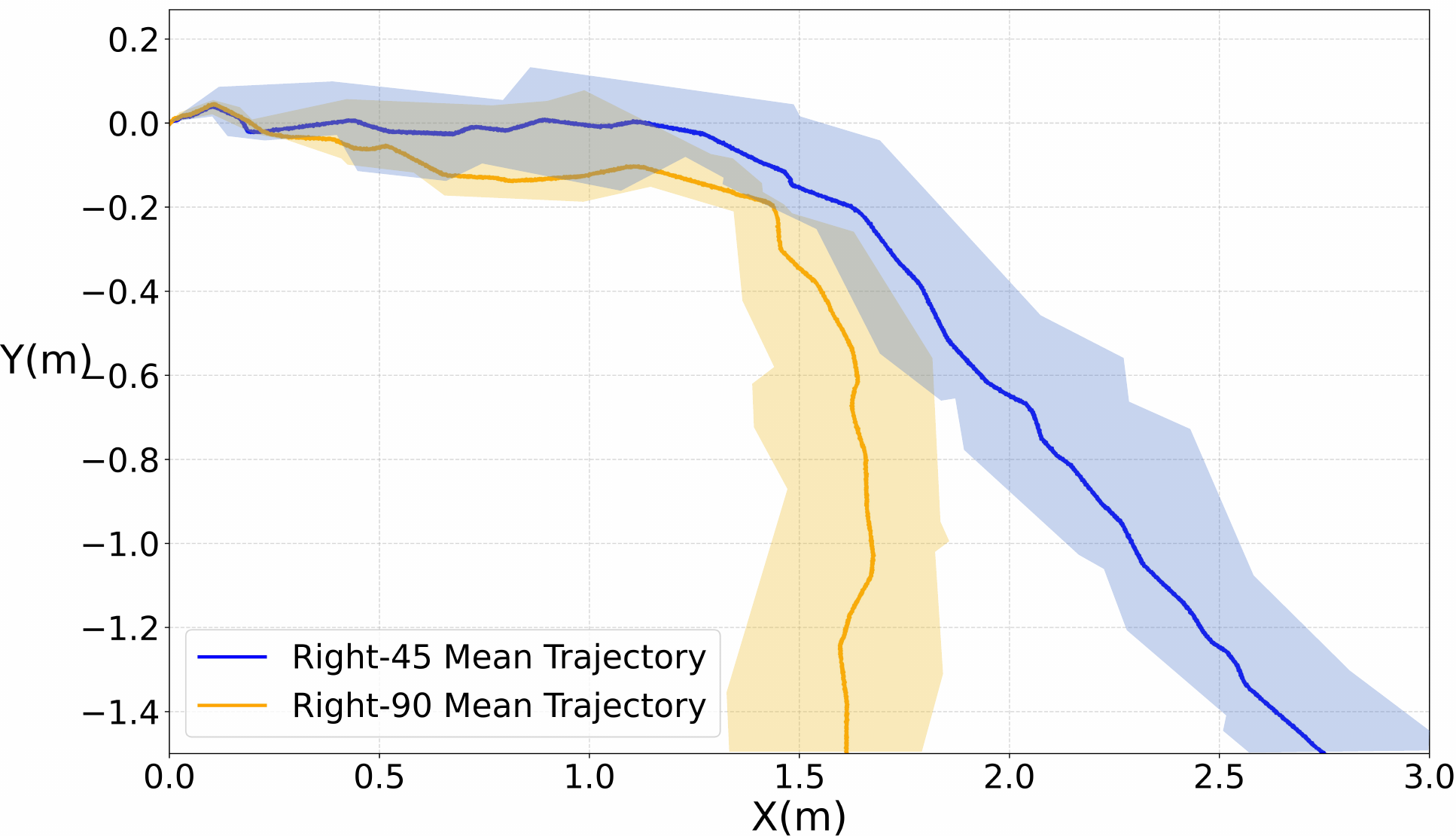}
        \caption{Right-turn dribbling trajectory}
    \end{subfigure}
    \caption{Mean ball trajectories (solid blue and orange lines)  for four dribbling maneuvers. The blue and orange shaded regions represent the variation across all individual trajectories for the 45-degree turn and the 90-degree turn, respectively.}
    \label{fig:accuracy}
\end{figure}

To evaluate the velocity tracking accuracy of our proposed method, we conduct a simulation experiment in MuJoCo to record the trajectory of the ball during a dribbling task. The goal is to compare the actual ball velocity after turning left/right with the commanded target velocity, as shown in Fig.~\ref{fig:accuracy}. 
The ball is initially placed at (0, 0) and then is commanded to move straight along the positive x-axis. When the ball enters a predefined area---a circle with a radius of 0.4 meters centered at (1.5, 0)---the direction command is switched to a 45-degree and 90-degree turn, respectively, either to the left or right, introducing a sharp directional change. We conduct 5 roll-outs for each turn direction and each turning angle with ±0.1 m/s random command perturbations, resulting in a total of 20 trials. 

\subsubsection{Experiment Results}
To thoroughly evaluate the performance of our system, we divide the metrics into two parts:
\begin{itemize}
    \item \textbf{Direction Tracking Error}: The difference between the actual ball velocity direction change and the target ball velocity direction change. It measures how effectively the system can maintain alignment with the intended dribbling direction during sharp direction changes.
    \item \textbf{Speed Tracking Error}: The difference between the actual magnitude of ball velocity and the target magnitude of ball velocity. It reflects the system’s ability to accurately regulate and maintain the desired ball speed during dribbling.
\end{itemize}

For direction tracking effectiveness, our method demonstrates consistent and accurate alignment with the target direction across all trials. Fig. \ref{fig:accuracy} overlays the trajectories in simulation for both left and right turns, illustrating that the ball remains closely aligned with the target direction. We define direction tracking error as the difference between the trajectory direction change and the target direction change, which serves as a key metric for evaluating direction tracking. Specifically, the direction change of each trajectory is calculated by fitting two straight lines to the trajectory points before and after the turn and computing the angular difference between them, while the target direction change is set to 45 degrees or 90 degrees. Under the left-turn dribbling task, the average direction change is 43.58 degrees for the 45-degree target, corresponding to a relative error of 3.16\%, and 88.06 degrees for the 90-degree target, corresponding to a relative error of 2.16\%. Under the right-turn task, the average direction change is 46.44 degrees for the 45-degree target, with a relative error of 3.20\%, and 87.98 degrees for the 90-degree target, with a relative error of 2.24\%. These results highlight our method's capability to respond effectively to abrupt directional changes while maintaining precise control of the dynamic system.

For speed tracking performance, we evaluate system performance by calculating the ball speed error. The average ball speed for each trajectory is computed and compared to the target speed of 1 m/s. Results indicate that, during left-turn dribbling, the average ball speed of all trials is 0.898 m/s, with a relative error of 10.3\%. For right-turn dribbling, these results are 0.896 m/s and 10.4\%, respectively. These findings suggest that the system achieves accurate speed tracking.

\subsection{Ablation Study in Real World}
\subsubsection{Experiment Setup}
To assess the contributions of each component in our approach, we conduct an ablation study comparing our full method against the following baselines: (1) dribbling with direct joystick tele-operation using the company's default locomotion policy, (2) our method without active sensing, and (3) single-stage training. 

For the direct tele-operation baseline, the operator employs a joystick to send base velocity commands directly to the robot, while the robot tracks it with the manufacturer-provided walking policy. In the second baseline, we retrain our method with the active sensing module disabled. For the third baseline, we use only single-stage training instead of our two-stage setup. For our method, the second and third baselines, the operator provides target ball velocity commands, which are used to indirectly control the robot.

\begin{figure}[h]
    \centering
    \includegraphics[width=\linewidth]{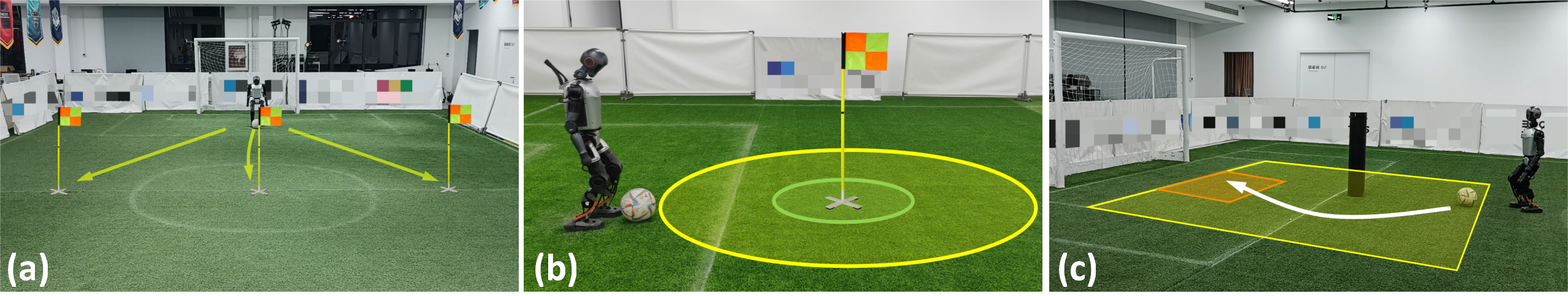}
    \caption{\textbf{Ablation study tasks setup.} \textbf{(a)(b)} The three target positions and the region definition of success and failure in the Dribbling to Target task. \textbf{(c)} The region setup in the Obstacle Avoidance task.}
    \label{fig:task_pic}
\end{figure}

We evaluate our method and all the baselines on two tasks: Dribbling to Target and Obstacle Avoidance, as shown in Fig.~\ref{fig:task_pic}. For Dribbling to Target, the operator controls the robot to dribble the ball to targets. A trial is considered successful if the final position error is less than 1 meter. Failures occur if the ball ever enters the 3 m radius around the target (yellow circular region), fails to reach the 1 m success zone (green circular region), and subsequently exits the 3 m radius, or if the ball leaves the boundary of the soccer field at any time. For Obstacle Avoidance, the operator uses the joystick to control the system to dribble while avoiding a single obstacle. Success requires reaching the target region (orange box) without the robot or ball colliding with the obstacle or leaving the task region (yellow box). 15 trials are conducted for both tasks. We record the success rate and the average completion time of all trials.

\subsubsection{Experiment Results}
We evaluate the performance of our method against several baselines on the two tasks. The results are summarized in Table \ref{table:results}.

For the Dribbling to Target task, our method attains an 86.7\% success rate with an average time usage of 22.0 seconds. This significantly outperforms direct tele-operation, which achieves a 66.7\% success rate but requires almost double the time (42.3 seconds). This is the result of not controlling the velocity of the ball directly and the operator has to spend more effort adjusting the robot to align with the ball and then move forward. The force of kicking is difficult to control and the robot could easily kick the ball further away than needed. Moreover, the method without active sensing and the one-stage training approach perform notably worse, with success rates of 13.3\% and 33.0\%, respectively. Interestingly, while the method without active sensing completes the task in the same amount of time (22.0 seconds) as our full method, its extremely low success rate suggests that active sensing is crucial for effective and reliable task execution. This is because the time usage is short if the first few kicks complete the task; however, if these initial attempts are unsuccessful, the lack of active sensing causes the robot to lose track of the ball, leading to failure.

In the Obstacle Avoidance task, our method achieves a 92.3\% success rate with an average completion time of 31.4 seconds. In contrast, direct tele-operation achieves only a 61.5\% success rate with a shorter average time of 23.0 seconds. The higher failure rate is attributed to the unpredictability of the ball's trajectory under manual control—without feedback-based regulation of the ball's motion, the ball may be kicked either too far or directly into obstacles. 
The operator only controls the robot’s velocity rather than directly controlling the ball, leaving the contact forces between the robot and the ball uncontrolled.

Notably, both the method without active sensing and the one-stage training baseline fail in this task, each achieving a 0\% success rate. The Obstacle Avoidance task is more challenging, and the policy without active sensing can easily lose track of the ball, whereas our policy can actively search for the ball when it leaves the field of view. Also, the one-stage training policy struggles to walk stably while controlling the ball, since both skills must be learned jointly from scratch without the benefits of staged curriculum. These results underscore the necessity of incorporating both active sensing and multi-stage training for reliable dribbling. 

\subsection{Real-World Generalization Study}

\subsubsection{Experiment Setup}
We test the generalization abilities of our proposed method according to two aspects:

\begin{itemize}
    \item \textbf{Generalization to Various Humanoid Hardware}: Whether our proposed method can be deployed on different humanoid/bipedal platforms for dribbling.
    \item \textbf{Generalization to Diverse Terrains}: Whether Dribble Master can adapt to different real-world terrains and dribble smoothly.
\end{itemize}

To evaluate generalization abilities across different humanoid/bipedal robots, we deploy our trained policy on the High Torque Mini Pi bipedal robot and the Unitree G1 humanoid robot in simulation and command them to track the target ball velocity.

\begin{figure}[h]
    \centering
    \includegraphics[width=\linewidth]{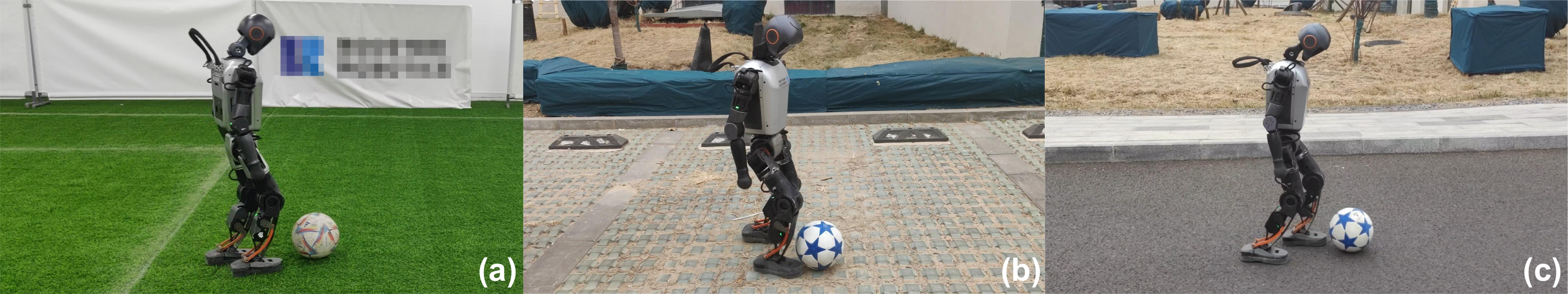}
    \caption{\textbf{Dribbling on diverse rough terrains.} \textbf{(a)} Artificial Grass, \textbf{(b)} Potholed Ground, \textbf{(c)} Asphalt Pavement.}
    \label{fig:terrain_pic}
\end{figure}

To evaluate the generalization capabilities on diverse terrains, we conduct dribbling experiments in diverse real-world environments, including a standard soccer field, an asphalt surface, and an uneven terrain with non-structured protrusions (e.g., gravel or small obstacles), as shown in Fig. \ref{fig:terrain_pic}. These terrains pose unique challenges due to variations in surface friction and irregular contact dynamics, which can destabilize both the robot's gait and the ball's motion.

On an asphalt surface, lower friction levels affect traction and stopping precision, while on irregular terrain, small bumps and protrusions introduce unpredictable perturbations to the ball trajectory and foot placement. 
In all scenarios, the robot is tasked with dribbling the ball along a predefined trajectory using the same learned policy, without environment-specific fine-tuning.

\begin{figure}[h]
    \centering
    \includegraphics[width=\linewidth]{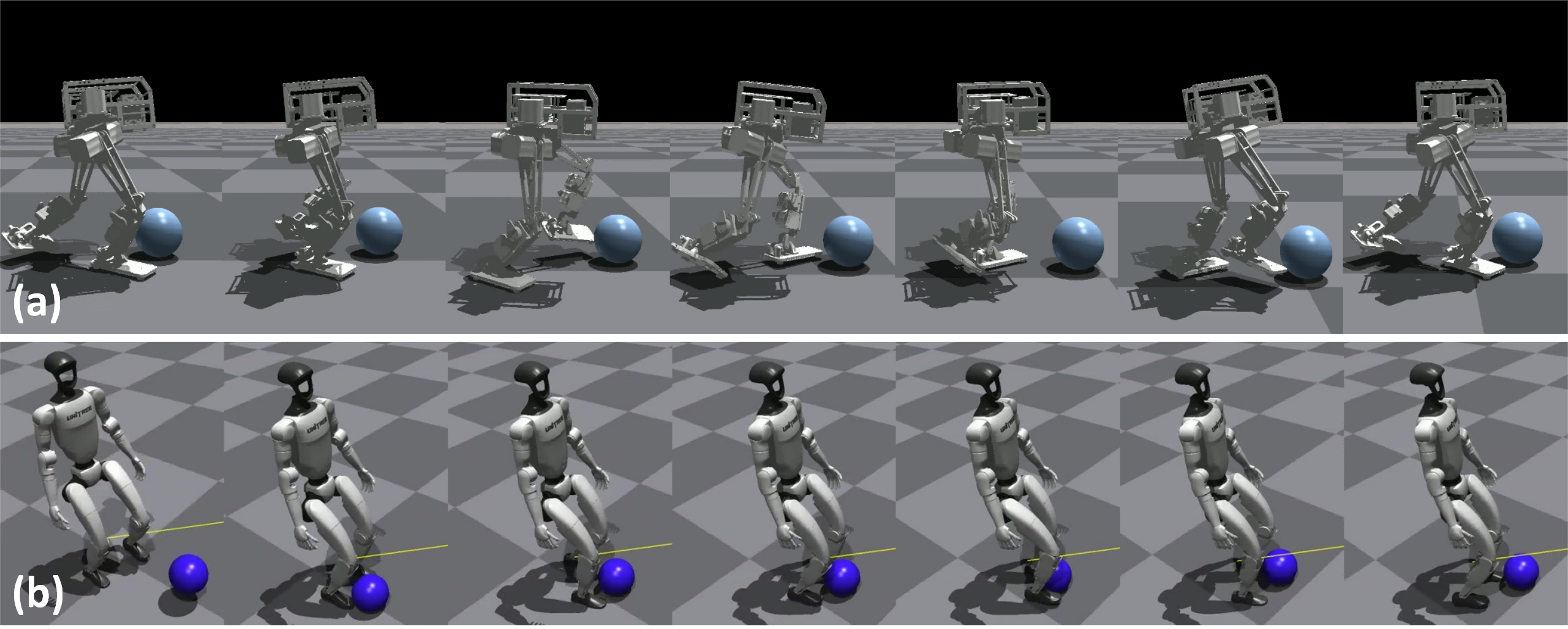}
    \caption{\textbf{Generalization to diverse robot platforms.} \textbf{(a)} High Torque Mini Pi bipedal robot, \textbf{(b)} Unitree G1 humanoid robot.}
    \label{fig:pi_g1_pic}
\end{figure}

\subsubsection{Experiment Results}

For generalization to different robots, the simulation results have shown that different robots can dribble well and perform dribbling motions with good visual appeal, as presented in Fig. \ref{fig:pi_g1_pic}. For generalization to diverse terrains, as shown in the accompanying video and Fig. \ref{fig:terrain_pic}, the robot successfully performs the dribbling task across all tested terrains. While some deviations from the reference trajectory are observed—particularly on the uneven ground—the system remains stable and consistently completes the task without external intervention.

Compared to prior works such as \cite{tirumala2024soccervision} and \cite{haarnoja2024learning}, which rely heavily on environment-specific setup and are especially sensitive to domain shifts, our approach achieves stronger real-world generalization with lower computational overhead. By isolating motion control from the visual front-end, we reduce the burden of retraining in new environments and improve robustness to terrain-induced disturbances and visual noise. These results affirm the practicality of our method for deploying robotic soccer systems in unstructured and dynamic outdoor settings.

\section{CONCLUSIONS AND FUTURE DIRECTIONS}
We propose a two-stage curriculum reinforcement learning framework for humanoid dribbling, introducing a virtual camera in simulation to address discontinuous ball perception and enable active sensing. We successfully transfer our trained policy to the real world, which has shown that our method enables agile dribbling skills. Generalization experiments demonstrate robust performance across diverse terrains and successful deployment on different robot platforms.

For future work, we plan to equip the robot with autonomous obstacle avoidance capabilities to enhance its navigational awareness. We also intend to extend our research to more complex situations featuring multiple moving humanoid robots and adversarial challenges, to achieve more dexterous dribbling and develop a strategic multi-agent coordination. Additionally, we will explore integrating imitation learning from demonstrations to further refine and enrich dribbling behaviors. We believe these improvements will lead to more dynamic and visually impressive results.

\section{ACKNOWLEDGMENT}
We sincerely thank Booster Robotics for hardware support.

\bibliographystyle{IEEEtran}
\bibliography{root}

\end{document}